\documentclass[11pt,a4paper]{article}
\usepackage[utf8]{inputenc}
\usepackage[T2A,T1]{fontenc}
\usepackage[russian,english]{babel}
\usepackage[hyperref]{emnlp-ijcnlp-2019}
\usepackage{times,graphicx}
\usepackage{floatrow}
\usepackage{latexsym,booktabs,multicol,multirow}
\usepackage{amssymb}
\usepackage{pifont}
\newcommand{\cmark}{\ding{51}}%
\newcommand{\xmark}{\ding{55}}%
\usepackage{url}
\usepackage{textcomp}
\usepackage{color, colortbl}
\usepackage{caption}
\usepackage{adjustbox}
\usepackage{multirow}

\usepackage{url}

\aclfinalcopy 

\title{Lost in Evaluation:\\ Misleading Benchmarks for Bilingual Dictionary Induction}

\author{Yova Kementchedjhieva \\
  University of Copenhagen\\
  {\tt yova@di.ku.dk} \\\And
  Mareike Hartmann \\
  University of Copenhagen\\
  {\tt hartmann@di.ku.dk} \\\And
  Anders Søgaard \\
  University of Copenhagen\\
  {\tt soegaard@di.ku.dk} \\}

\date{}

\begin{document}
\maketitle

\begin{abstract}
The task of bilingual dictionary induction (BDI) is commonly used for intrinsic evaluation of cross-lingual word embeddings. The largest dataset for BDI was generated automatically, so its quality is dubious. 
We study the composition and quality of the test sets for five diverse languages from this dataset, with concerning findings: (1) a quarter of the data consists of proper nouns, which can be hardly indicative of BDI performance, and (2) there are pervasive gaps in the gold-standard targets. 
These issues appear to affect the ranking between cross-lingual embedding systems on individual languages, and the overall degree to which the systems differ in performance. With proper nouns removed from the data, the margin between the top two systems included in the study grows from 3.4\% to 17.2\%. Manual verification of the predictions, on the other hand, reveals that gaps in the gold standard targets artificially inflate the margin between the two systems on English to Bulgarian BDI from 0.1\% to 6.7\%. 
We thus suggest that future research either avoids drawing conclusions from quantitative results on this BDI dataset, or accompanies such evaluation with rigorous error analysis.
\end{abstract}

\section{Introduction}
Bilingual dictionary induction (BDI) refers to retrieving translations of individual words. 
The task has been widely used for intrinsic evaluation of cross-lingual embedding algorithms, which aim to map two languages into the same embedding space, for transfer learning purposes \cite{C12-1089}. Recently, \citet{Glavas19} reported limited evidence in support of this practice---they found that cross-lingual embeddings optimized for a BDI evaluation metric were not necessarily better on downstream tasks. 
Here, we study BDI evaluation in itself, as has been done for other evaluation methods in the past (cf. \citealp{faruqui2016problems}'s work on word similarity), with concerning findings about its reliability. 

A massive dataset of 110 bilingual dictionaries, known as the \texttt{MUSE} dataset, was introduced in early 2018 along with a strong baseline \cite{Conneau2018}. Subsets of the \texttt{MUSE} dictionaries have been used for model comparison in the evaluation of numerous cross-lingual embedding systems developed since (cf. \citealp{grave2018unsupervised,jawanpuria2019, Hoshen2018AnIC, hoshen2018non, DBLP:journals/corr/abs-1809-02306,joulin2018rcsls}). Even though the field has been very active, progress has been incremental for most language pairs.
Moreover, there have been very few attempts at a linguistically-informed error analysis of BDI performance as measured on \texttt{MUSE} (cf. \citealp{kementchedjhieva2018generalizing}). This is problematic for two reasons: on one hand, most systems greatly vary in their approach and architecture, so it is difficult to identify the source of the reported performance gains; on the other hand, the \texttt{MUSE} dataset was compiled automatically, with no manual post-processing to clean up noise, so the real impact of the performance gains is unclear.

In this work, we study the composition and quality of the \texttt{MUSE} data for five diverse languages:  German, Danish, Bulgarian, Arabic and Hindi. A manual part-of-speech annotation of the test sets for these languages 
reveals a strikingly high number of proper nouns. We refer to linguistic literature to argue that proper nouns,
having no lexical meaning but rather just a referential function, cannot reliably be used in the evaluation of word-level translation systems.
We find that excluding proper noun pairs from the test dictionaries for the aforementioned languages affects the ranking and degree of performance gaps between five of the most influential recent systems for BDI.

With a new, more reliable ranking at hand, we perform qualitative analysis on the performance gap between the best and second best systems for Bulgarian. This reveals another major issue with the data: limited coverage of morphological variants for the target words. Through manual verification of the models' predictions, we find that the gap in performance between the two systems is far smaller than previously perceived. 

The uncovered issues of high noise levels (proper nouns) and limited coverage (missing gold standard targets) clearly have a crucial impact on BDI results obtained on the \texttt{MUSE} dataset, and need to be addressed. Filtering out proper nouns could be achieved automatically, by checking against gazetteers of named entities. We find that an automatic procedure for the filling of missing targets, however, yields only minor improvements. We thus urge researchers to be cautious when reporting quantitative results on \texttt{MUSE}, and to account for the problems presented here through manual verification and analysis of the results. As an alternative, we point them to morphologically complete BDI resources, built bottom-up \cite{czarnowska2019dont}. We share our part-of-speech annotations, such that future work can use this resource for analysis purposes. \footnote{Available at https://github.com/coastalcph/MUSE\_dicos}

\section{Bilingual Dictionary Induction}\label{sec:systems}

Improvements on BDI mostly stem from developments in the space of cross-lingual embeddings, which use BDI for intrinsic evaluation.

\paragraph{Systems} Five influential recent systems for cross-lingual embeddings are MUSE \cite{Conneau2018}, which can be supervised (\textbf{MUSE-S}) or unsupervised (\textbf{MUSE-U}); VecMap, which also can be supervised (\textbf{VM-S}) \cite{Artetxe2018} or unsupervised (\textbf{VM-U}) \cite{Artetxe2018b}; and RCSLS \cite{joulin2018rcsls}, a supervised system (\textbf{RCSLS}), which scores best on BDI out of the five. We refer the reader to the respective publications for a general description of the systems.

\paragraph{Metrics} Performance on BDI in these works is evaluated by verifying the system-retrieved translations for a source word against a set of gold-standard targets. The metric used is Precision at $k$ (P@$k$), which measures how often the set of $k$ top predictions contains one of the gold-standard targets, i.e. what is the ratio of True Positives to the sum of True Positives and False Positives.

\paragraph{Data} All systems listed above report results on one or both of two test sets: the \texttt{MUSE} test sets \citet{Conneau2018} and/or the \texttt{Dinu} test sets \cite{Dinu2015, Artetxe2017}. Similarly to \texttt{MUSE}, the \texttt{Dinu} dataset was compiled automatically (from Europarl word-alignments), but it only covers four languages. Due to the bigger size of \texttt{MUSE} (110 language pairs), we deem its impact larger and focus our study entirely on it. 

\addtolength{\textfloatsep}{-0.1in}

\section{Annotation-based observations}

In order to gain insights into the linguistic composition of the \texttt{MUSE} dictionaries, we employ annotators fluent in German, Danish, Bulgarian, Arabic and Hindi (hereafter, {\sc de, da, bg, ar, hi}) to annotate the entire dictionaries from English to one of these languages (hereafter, from-{\sc en}) and the entire dictionaries from these languages to English (hereafter, to-{\sc en}) with part-of-speech (POS) tags. Details on the annotation procedure can be found in Appendix A. Below, we discuss our findings on the POS composition of the data,
and we evaluate the performance of RCSLS per POS tag.\footnote{For all experiments, we use the pretrained embeddings of \citet{bojanowski2017enriching}, trained on Wikipedia.}

\subsection{Analysis of POS composition}

The average percentage of common nouns, proper nouns, verbs, and adjectives/adverbs in the dictionaries to-{\sc en} was respectively 49.6, 24.9, 12.5, and 12.9.\footnote{The numbers were similar across from-{\sc en} dictionaries.} Nouns constitute half of the dictionaries' volume, while verbs and adjectives/adverbs collectively make up only about a fourth of the average dictionary. A skewed ratio between these three categories is not surprising: in the EWT dependency treebank, for example, which contain gold-standard POS tags, the proportion of noun, verb and adjective/adverb types is 34, 17 and 14 percent, respectively. Notice, however, that in the case of the \texttt{MUSE} data, the ratio is even more skewed in favour of nouns over the other two categories. 

The large number of proper nouns in the dictionaries seems even more problematic. Proper nouns are considered to have no lexical meaning, but rather just a referential function \cite{pierini}. Personal names usually refer to a specific referent in a given context, but they can, in general, be attributed to different referents across different contexts, and they are almost universally interchangeable in any given context.
Some personal names and most place and organization names may have a unique referent, e.g. \textit{Barack Obama, Wisconsin, Skype}, but these names still do not carry a \textit{sense}, their referent is resolved through access to encyclopedic knowledge \cite{pierini}. Considering that the pretrained embeddings which we use were trained on Wikipedia, we can expect that such encyclopedic information would indeed appear in the context of certain unique names, but importantly, the alignability of the embeddings for such entities would depend on the level of parallelism between the contents of Wikipedia articles in the different languages.

With these considerations in mind, one should wonder how stable the representation of names can be in an embedding space. This question has previously been raised by \citet{Artetxe2017}. We address it empirically below.

\subsection{Evaluation by POS}

\begin{figure}
    \resizebox{\linewidth}{!}{
    \centering
    \includegraphics{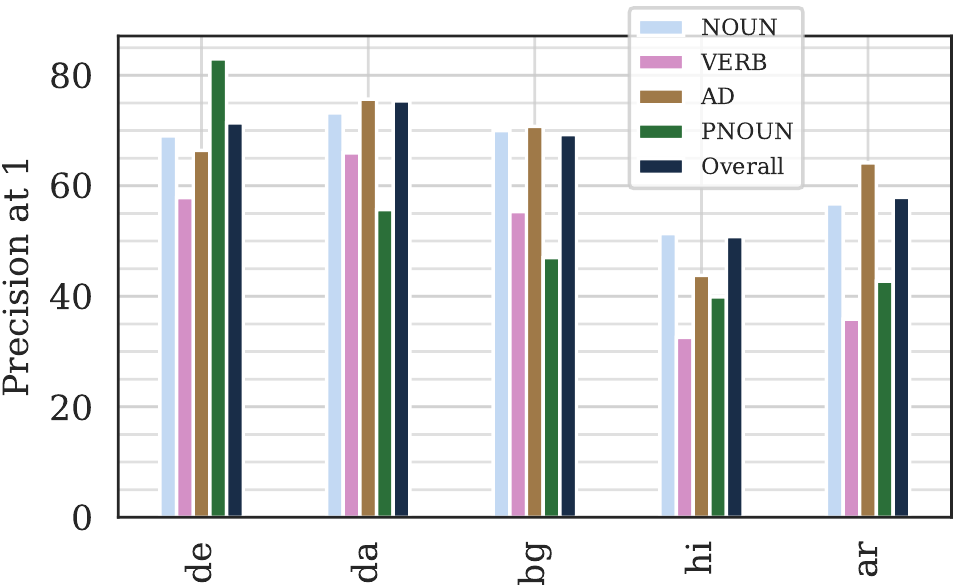}
    }
    \caption{Precision of RCSLS by POS tag on to-{\sc en} data.}
    \label{fig:pos}
\end{figure}

Figure~\ref{fig:pos} shows the precision of the RCSLS embedding alignment method on different POS segments of the test data in mapping to-{\sc en} (results from-{\sc en} were similar and are shown in Appendix B). Verbs pose a greater challenge to BDI systems than nouns and adjectives do. 
Generally, we can attribute this observation to the higher abstraction of concepts described by verbs. This is a known problem for word embedding methods in general \cite{Gerz2016}, which BDI systems naturally inherit. 

With respect to proper nouns, we observe that they indeed introduce a level of instability in the evaluation of BDI systems. Notice that while the other parts of speech follow a similar pattern across languages, with higher precision obtained for nouns and adjectives/adverbs than for verbs, relative precision on proper nouns is highly variable. For {\sc de}, proper nouns are easier to translate than other parts of speech by a margin of 15\%, for {\sc hi} and {\sc ar} they are easier than nouns and adjectives/adverbs, but harder than verbs, and for {\sc da} and {\sc bg} they are hardest out of all four categories. We looked into the individual word pairs marked as proper nouns in the {\sc de} and {\sc da} data, as these languages are related and RCSLS performs comparably on them otherwise, and did not find any patterns that could explain the large differences. In fact, between the 384 proper noun pairs in the {\sc en-de } dictionary and the 330 proper noun pairs in the {\sc en-da} dictionary, there was an overlap of 279 pairs, retrieved with precision of 89.21\% in the {\sc en-de } setting and 51.30\% in the { \sc en-da } setting. We conjecture that this result relates to the level of parallel content between the Wikipedia dumps for the different language pairs, which is likely higher for {\sc en-de }, since the dumps for these languages are also closer in size: 5.8M articles in {\sc en}, 2.3M in {\sc de} (and only 0.2M in {\sc da}).\footnote{https://meta.wikimedia.org/wiki/List\_of\_Wikipedias} 

We evaluate this hypothesis through an experiment where we train an RCSLS alignment for {\sc de-en} using the {\sc de} embeddings of \citet{Artetxe2017}, trained on SdeWaC \cite{baroni2009wacky} and the {\sc en} embeddings of \citet{Dinu2015}, trained on ukWaC \cite{baroni2009wacky}, Wikipedia and the BNC \footnote{Available at http://www.natcorp.ox.ac.uk} corpora. The level of parallel content between the data used to train the two sets of embeddings is thus far more limited in this case, and the {\sc de} embeddings are not explicitly trained on Wikipedia data. Table~\ref{tab:wacky} summarizes the results: while with the new embeddings performance is somewhat reduced for nouns, verbs and adjectives/adverbs, precision at 1 for proper nouns, in particular, drops by over 50\%, indicating that this category of test word pairs is indeed highly sensitive to the nature of the training data. 

\begin{table}
\resizebox{\linewidth}{!}{
\begin{tabular}{llllll}
\toprule
Corpora&{\sc noun}&{\sc verb}&{\sc ad}&{\sc pnoun}\\
\midrule
Wikipedia&69.0&57.9&66.4&83.0 \\
Mixed*&64.0&55.5&59.4&37.6 \\
\bottomrule
\end{tabular}
}
\caption{Comparison in performance by POS category with two different embedding sets. * The out-of-vocabulary rate for items in the dictionaries is negligible: 2, 0, and 1 for {\sc noun}, {\sc verb }, and {\sc ad} , respectively.}
\label{tab:wacky}
\end{table}

\begin{figure}
    \centering
    \resizebox{\linewidth}{!}{
    \includegraphics{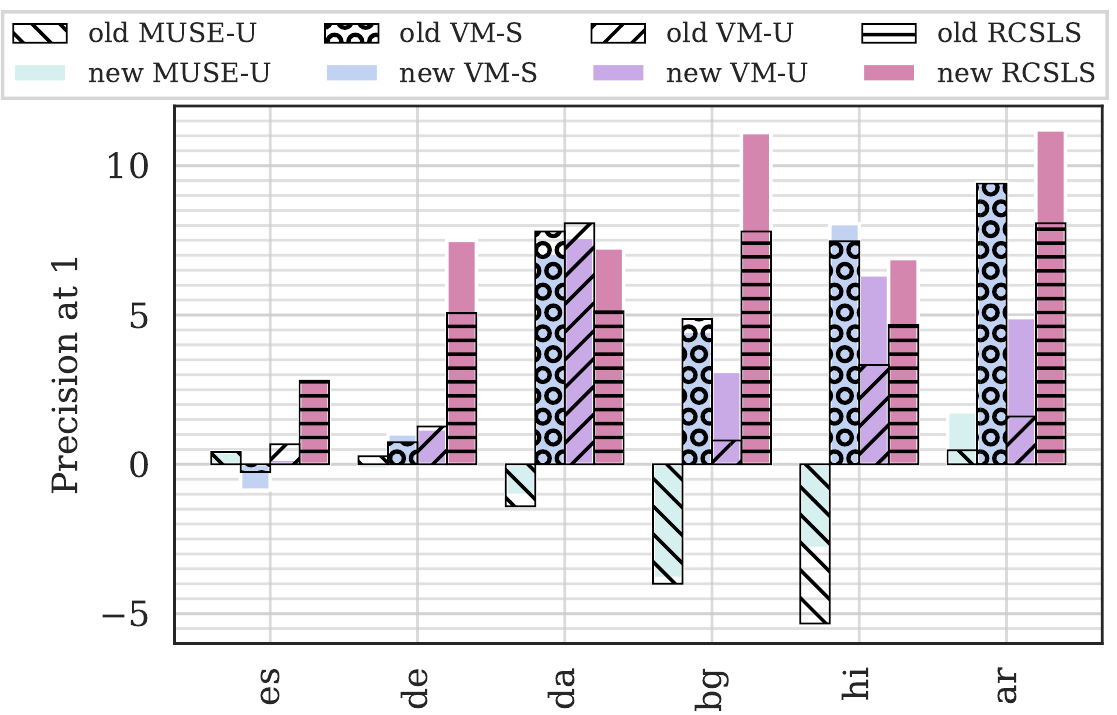}%
    }
    \caption{Absolute difference in performance on from-{\sc en} BDI, relative to MUSE-S. Pattern-filled bars show results as estimated on the original data (\textit{old}), while colored bars show results as estimated on the cleaned data (\textit{new}).}
    \label{fig:my_label}
\end{figure}

\subsection{Re-ranking on clean data}

Based on the analysis presented above, we removed all pairs that were annotated as proper nouns and all pairs that were marked as invalid during the annotation process.\footnote{The latter constitute less than 1\% of the removed data.} This clean-up resulted in a drop in the size of the test dictionaries of about 25\% on average.
A detailed size comparison between the old test dictionaries and their new cleaned versions is presented in the top rows of Table~\ref{tab:my_label} in Appendix B. Figure~\ref{fig:my_label} visualizes a re-evaluation of the five systems for BDI listed in Section~\ref{sec:systems}, on the original test data and on the new clean versions of the test dictionaries from-{\sc en}.\footnote{The to-{\sc en} results were similar, see Appendix B.} The results are reported in terms of change in performance relative to MUSE-S (chosen as a baseline) as estimated on the original \texttt{MUSE} data (pattern-filled bars) and on the cleaned version of the data (colored bars). The absolute system performances before and after the clean-up can be found in Table~\ref{tab:my_label} in Appendix B. 

We see that the ranking between the models changes most notably for {\sc ar}, where RCSLS appears inferior to VM-S on the original test data, but on the clean data it emerges as best. For {\sc bg}, the evaluation on the clean test data reveals that RCSLS outperforms the next best system, VM-S, by a larger factor than it appeared on the original test data. Lastly, for {\sc da}, evaluation on the original test data makes RCSLS seem far inferior to VM-S and VM-U, but on the clean test data we see that it outperforms VM-S and matches the performance of VM-U. These observations show that the noise coming from proper nouns has a large impact on the perceived ranking and difference in performance between systems. 

\section{\textit{False} False Positives}\label{ffp}
With a more reliable estimate of the models' performance at hand, we next manually study the remaining performance gap between RCSLS, the best-performing model overall, and VM-S, the second best model overall, for {\sc en--\sc bg}.\footnote{We also analyzed {\sc en--\sc de}, with very similar results.} We present some examples in Table~\ref{tab:examples_small} and more can be found in Table~\ref{tab:examples}, Appendix C. 

\begin{table*}
\centering
\begin{adjustbox}{width=\textwidth}
\begin{tabular}{c|lllll}
\toprule
Ex.&{\sc SRC}&{\sc TGT}&{\sc RCSLS}&{\sc VM-S}&Description\\
\midrule
A&joke& \foreignlanguage{russian}{шега}&\underline{\foreignlanguage{russian}{шега}}
[INDEF]&\foreignlanguage{russian}{шегата} [DEF] &definite form missing from targets\\
& & \foreignlanguage{russian}{лаф} \\
& & \foreignlanguage{russian}{виц} \\
B&remembered& \foreignlanguage{russian}{запомнен}&\underline{\foreignlanguage{russian}{запомнен}}[MASC]&\foreignlanguage{russian}{запомнена}[FEM]&feminine form missing from targets\\
C&hide& \foreignlanguage{russian}{скриване}&\foreignlanguage{russian}{скриване} [NOUN]&\foreignlanguage{russian}{скриват}[VERB]&\textit{hide} as a verb vs. \textit{hide} as a noun\\
D&bench& \foreignlanguage{russian}{пейка}&\foreignlanguage{russian}{пейка}&\foreignlanguage{russian}{скамейка}&synonym missing from targets\\
& & \foreignlanguage{russian}{пейката} \\
E&depot& \foreignlanguage{russian}{депо}&\foreignlanguage{russian}{депо}&\textcolor{gray}{\foreignlanguage{russian}{гара}}&VM-S predicted `train station'\\
F&crowned& \foreignlanguage{russian}{коронован}&\foreignlanguage{russian}{коронована}[FEM]&\underline{\foreignlanguage{russian}{коронован}}[MASC]&feminine form missing from targets\\
G&pond& \foreignlanguage{russian}{езерце}&\textcolor{gray}{\foreignlanguage{russian}{къщичка}}&\foreignlanguage{russian}{езерце}&RCSLS predicted `cottage'\\
H&grants& \foreignlanguage{russian}{субсидии}&\foreignlanguage{russian}{стипендии}&\foreignlanguage{russian}{стипендии}&synonym missing from targets\\
I&armies& \foreignlanguage{russian}{армии}&\foreignlanguage{russian}{армиите}&\foreignlanguage{russian}{армиите}&definite form missing from targets\\
\bottomrule
\end{tabular}
\end{adjustbox}
\caption{Example translations from {\sc en} to {\sc bg}. Underlined forms are more canonical. Grey forms are incorrect.}
\label{tab:examples_small}
\end{table*}

We find that there are 125 source words that RCSLS translated correctly and VM-S did not. Upon closer inspection, we find that for 54\% of these words, both RCSLS and VM-S predicted a valid translation, but RCSLS predicted a more \textit{canonical} translation, which was listed among the gold-standard targets, while VM-S predicted another word form that was missing from the list of gold-standard targets. By more canonical we mean, for example, indefinite instead of definite forms of nouns and adjectives (see Ex. A, Table~\ref{tab:examples_small}, masculine instead of feminine or neuter forms of adjectives (see Ex. B), singular instead of plural forms.
To the extent that a more canonical translation should be considered better, RCSLS is definitely showing superiority over VM-S. It is not clear, however, if that should be the case, since for some words, the test dictionary exhibits higher coverage than for others, i.e. the less canonical translations are not omitted by design, but appear to be accidental gaps. 

Another 19\% of the instances where RCSLS outperformed VM-S, we find to be clear cases of a missing translation in the test dictionary, i.e. not a missing form of a listed target, but a missing synonym or a missing sense altogether (see Ex. C and D).

The two types of errors in precision at 1 discussed above can be considered cases of \textit{false} False Positives, because they really should have been True Positives.
The remaining 27\% of the gap between the two models' performance indeed illustrate that RCSLS provides better translations in some cases (see Ex. E). 

Notice, however, that it is not the case that RCSLS outperformed VM-S in all cases--for 50 test words, VM-S predicted a correct translation and RCSLS did not. Among these, there are cases of missing translations from the dictionary as well (see Ex. F), but they can explain less of the lack in performance of RCSLS, i.e. 50\% of the translations of RCSLS are indeed erroneous (see Ex. G). 

To summarize, originally the performance gap between the two models appeared to be $(125-50)/1125*100=6.67\%$, while after the manual verification, it is $(27\%*125-50\%*50)/1125*100=0.1\%$.\footnote{1125 is the total dictionary size.} Such a substantial narrowing in the gap between the two models clearly indicates that conclusions drawn on the original result, i.e. that RCSLS is far superior that VM-S for this language pair, is hardly supported by the updated result.

A surface analysis of the subset of words for which neither RCSLS nor VM-S retrieved correct translations revealed similar patterns of extensive false False Positives, due to gaps in the coverage of the dictionary (see Ex. H and I). Our takeaway from these observations is two-fold. Firstly, when RCSLS retrieves a correct target form, it also usually retrieves its most \textit{canonical} form. More importantly, the evaluation of BDI systems on even the cleaned test dictionaries still does not represent accurately the differences in quality between them, due to major gaps in the coverage of the test dictionaries. 

\section{Concluding remarks}

Our study of the \texttt{MUSE} dataset revealed two striking problems: a high level of noise coming from proper nouns, and an issue of \textit{false} False Positives, due to gaps in the gold-standard targets. The former problem, we conjecture, can be solved by filtering names out with gazetteers. The quality of this solution would depend on the coverage of the gazetteers. 
The more challenging problem, however, is filling in the gaps, especially in terms of inflectional forms. We carried out preliminary experiments aiming to enrich the {\sc en--bg} and {\sc en-de} dictionaries. We extracted additional word forms of verbal and nominal targets from the UniMorph inflectional tables \cite{kirov2018unimorph}, according to a manually designed morphosyntactic correspondence map.\footnote{Details can be found in Appendix D.} Unfortunately, due to limited coverage of the UniMorph data, and, in the case of {\sc bg}, limited vocabulary of the pretrained embeddings, the impact of this procedure was almost negligible. Alternative approaches for enrichment exists, of course, but we wonder how worthwhile further efforts would be. That is, especially in light of \citealt{Glavas19}'s findings that BDI performance is not necessarily indicative of cross-lingual embedding quality. We therefore hope that our work adds weight to the call of \citet{Glavas19} for more reliable evaluation methods in cross-lingual embedding research. When BDI performance is used for evaluation purposes, it should be accompanied by manual verification, of the type presented here. 

\section{Acknowledgements}
We thank Marcel Bollmann, Matthew Lamm, Maria Barrett, Meriem Beloucif, Mostafa Abdou, Rahul Aralikatte and Victor Petrén Hansen for help with annotations. We would also like to thank Adam Lopez, Andreas Grivas and Sameer Bansal for useful feedback on drafts of the paper, and to the anonymous reviewers for their comments and suggestions. Anders S{\o}gaard was supported by a Google Focused Research Award; Mareike Hartmann by the Carlsberg Foundation. 

\bibliography{emnlp-ijcnlp-2019}
\bibliographystyle{acl_natbib_nourl}

\appendix

\newpage

\section{Appendix}

In order to obtain a reliable part-of-speech (POS) tagging of the \texttt{MUSE} test dictionaries efficiently, we used a two-step procedure. First, we ran the Stanford POS tagger \citep{Toutanova:2003} on the English side of each dictionary. We reduced the annotation schema to five categories: nouns (NOUN), proper nouns (PNOUN), verbs (VERB), adjectives and adverbs combined (AD), and others. Next, we asked NLP researchers with the appropriate language background to verify and correct the generated tags, based on both words in a pair. Where one word in the pair is ambiguous with respect to POS, but the other is not, they were told them to use the tag of the latter. If both words were ambiguous, we told them to use the tag they considered more frequent for these words. 

We instructed annotators that if a word can be both a proper noun and a common noun, it should be marked as the latter. We told them to mark pairs of identical words as proper nouns, under the assumption that they can be part of a company name or a brand, for example. That is, unless the words in the pair are actual cognates between the source and target language, or they are loanwords. See Table~\ref{tab:annot} for some examples. Lastly, we asked the annotators to mark pairs as invalid, if the source word is not a valid word in either the source or the target language, or the target word is not a valid translation of the source word. We note that this was a considerable annotation effort if over 40 hours in total. Each annotator had to process over 2000 word pairs: the dictionaries each consist of 1,500 source words, many of which have multiple translations, each processed separately. Annotation was performed in Microsoft Excel.

\begin{table}[!hb]
    \centering
    \resizebox{\textwidth}{!}{
    \begin{tabular}{llccc}
         {\sc src} & {\sc tgt} & {\sc pos} & valid & explanation  \\
         \midrule
         tea     & té       & {\sc noun} & \cmark & actual translation\\
         tea     & tea      & {\sc pnoun} & \cmark & part of a name,\\
                 &          &       &  & e.g. ``Lipton Iced Tea''\\
         rugby   & rugby    & {\sc noun}  & \cmark & loanword \\
         ugby    & ugby     & --    & \xmark & not a word in either language
    \end{tabular}
    }
    \caption{Example of annotated gold-standard word pairs from English to Spanish.}
    \label{tab:annot}
\end{table}

\newpage

\section{Appendix}

\begin{figure}[!hb]
    \resizebox{\linewidth}{!}{
    \centering
    \includegraphics{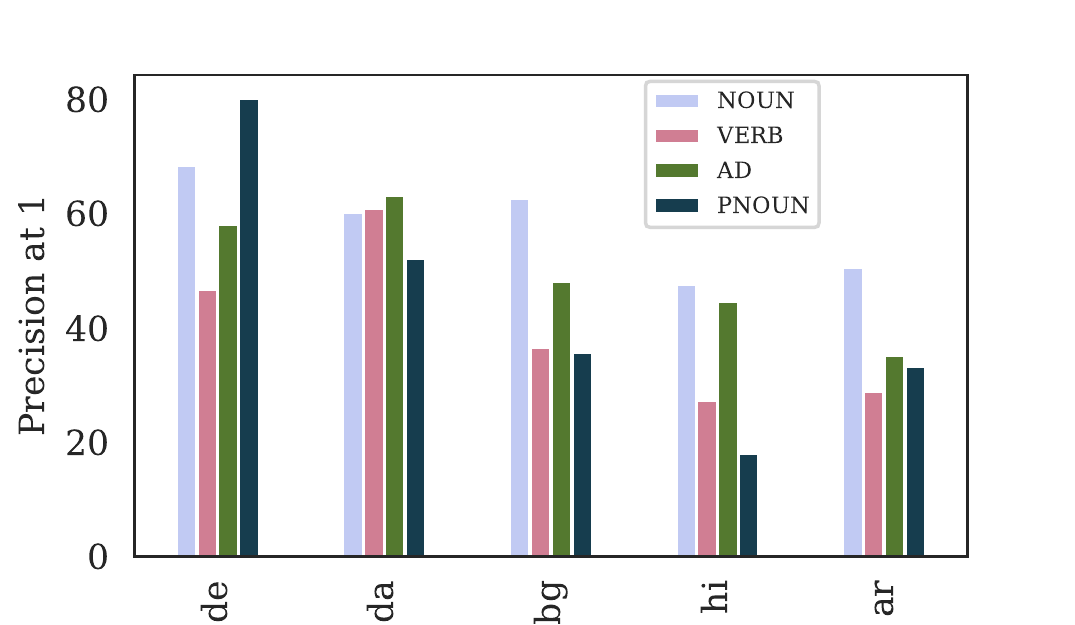}
    }
    \caption{Precision of the RCSLS system,  measured per POS tag, on to-{\sc en} data.}
    \label{fig:pos_appendix}
\end{figure}

The pattern of performance per POS tag is similar for to-{\sc en} mappings (see Figure~\ref{fig:pos_appendix}), as we saw it for from-{\sc en} mapping---proper nouns yield highly variable performance. 

\begin{figure}[!h]
    \centering
    \resizebox{0.9\linewidth}{!}{
    \includegraphics{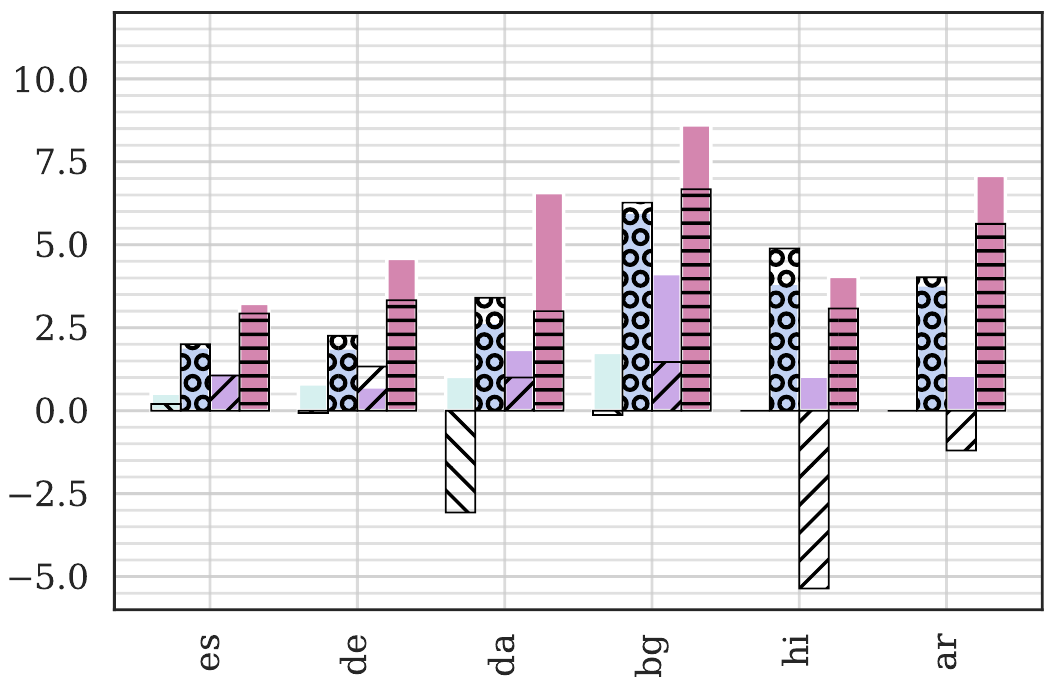}%
    }
    \caption{Change in performance on to-{\sc en} BDI relative to MUSE-S. Pattern-filled bars show results as estimated on the original data, while colored bars show results as estimated on the cleaned data.}
    \label{fig:my_label2}
\end{figure}

Similarly to mappings from-{\sc en}, in mappings to-{\sc en} (see Figure~\ref{fig:my_label2}) we see RCSLS outperforming other systems on the clean data for all languages (and by a large margin for most of them), whereas on the original data it appeared inferior to VM-S for {\sc da} and {\sc hi}. Another interesting observation here is that MUSE-U and VM-U occasionally appear inferior to the MUSE-S baseline (for {\sc da} and {\sc hi}, respectively) on the original test data, but on the clean test data all models yield an improvement over the baseline.\footnote{That is, excluding MUSE-U evaluated on {\sc hi} and {\sc ar}, where all solutions found were degenerate, so they have been excluded.}

\definecolor{LightCyan}{rgb}{0.5,1,1}
\begin{table*}
\resizebox{\textwidth}{!}{
\begin{tabular}{ccccccccccccc}
& \multicolumn{2}{c}{es}    & \multicolumn{2}{c}{de}    & \multicolumn{2}{c}{da}    & \multicolumn{2}{c}{bg}    & \multicolumn{2}{c}{hi}    & \multicolumn{2}{c}{ar} \\
& $\,\to\,$en & en$\,\to\,$ & $\,\to\,$en & en$\,\to\,$ & $\,\to\,$en & en$\,\to\,$  & $\,\to\,$en & en$\,\to\,$  & $\,\to\,$en & en$\,\to\,$  & $\,\to\,$en & en$\,\to\,$  \\
\rowcolor{LightCyan}
\multirow{2}{*}{Source words} & 1500 & 1500 & 1500 & 1500 & 1500 & 1500 & 1500 & 1500 & 1500 & 1500 & 1500 & 1500 \\
                            & 1145 & 1171 & 1111 & 1188 & 974 & 1158 & 1124 & 1125 & 963 & 1104 & 1212 & 1080 \\
\hline
\rowcolor{LightCyan}
\multirow{2}{*}{MUSE-S}   &  83.47 &  81.66 &  72.67 &  73.93  & 67.07 & 56.80 & 56.93  &  43.93  & 44.07  &  33.60   & 49.93 & 34.13  \\     
                          & 79.56  &  73.36 &  66.79 &  64.47  & 68.79 & 55.44  & 60.63 &  45.33  & 46.73  & 37.68    & 50.83 & 34.63  \\  
\rowcolor{LightCyan}
\multirow{2}{*}{MUSE-U}   & 83.67  &  82.07 &  72.60 &  74.20  & 64.00 & 55.40 & 56.80 &  39.93   & 0.00  &  28.27   & 0.00 & 34.60 \\
                          &  80.09 &  73.78 &  67.60 &  64.31  & 69.82 & 54.40 &  62.39 &  41.51   & 0.00  &  34.87 & 0.00 & 36.39 \\
\rowcolor{LightCyan}                       
\multirow{2}{*}{VM-S}     &  85.47 &  81.40 & 74.93 &  74.67  & 70.47 & 64.60 &  63.20 &  48.80 & 48.96  & 41.07  & 53.95 & 43.53 \\
                          &  81.48 &  72.50 &  68.68 &  65.49  & 71.46 & 62.52 & 66.61  & 49.78 & 50.57  & 45.74  & 54.62 & 44.07 \\
\rowcolor{LightCyan}     
\multirow{2}{*}{VM-U}     &  84.53 &  82.33 & 74.00  &  75.20  & 68.07 & 64.87 & 58.40 & 44.73 & 38.71 & 36.93 & 48.73 & 35.73  \\
                          &  80.70 &  73.53 & 67.51  & 65.66 & 70.64 & 63.04 & 64.76  & 48.44 & 47.77  &  44.02 & 51.90 & 39.54 \\
\rowcolor{LightCyan}
\multirow{2}{*}{RCSLS}    & 86.40 & 84.46 & 76.00 & 79.00 & 70.07 & 61.93 & 63.60 & 51.73 & 47.15 & 38.27 & 55.56 & 42.20 \\
                          & 82.79 & 76.17 & 71.38 & 71.97 & 75.36 & 62.69 & 69.24 & 56.44 & 50.78 & 44.57 & 57.92 & 45.83 \\
         
\end{tabular}
}

\caption{Cyan rows correspond to the original test data and white rows to the clean test data. The top rows report the sizes of the dictionaries, measured in terms of source words. For unstable models, e.g. MUSE-U, we train ten models and report results from one random successful model. For a fair comparison of MUSE-U and MUSE-S, we run Procrustes for 5 iterations in both cases, and use the same model selection criterion, mean cosine similarity, in both cases. All systems are evaluated using CSLS for retrieval. * Instead of full annotation for Spanish, we only mark proper nouns and remove them from the test dictionaries to and from English. }
\label{tab:my_label}
\end{table*}

\begin{table*}[hb!]
\centering
\begin{adjustbox}{width=\textwidth}
\begin{tabular}{c|lllll}
\multicolumn{2}{l}{\large \textbf{C \space\space\space Appendix}} \\
\multicolumn{2}{l}{} \\
\toprule
&{\sc SRC}&{\sc TGT}&{\sc RCSLS}&{\sc VM-S}&Description\\
\midrule
  \parbox[t]{2mm}{\multirow{13}{*}{\rotatebox[origin=c]{90}{VM-S \xmark, RCSLS \cmark }}}
&joke& \foreignlanguage{russian}{шега}&\underline{\foreignlanguage{russian}{шега}}&\foreignlanguage{russian}{шегата}&definite form missing from targets\\
& & \foreignlanguage{russian}{лаф} \\
& & \foreignlanguage{russian}{виц} \\
&arbitrators& \foreignlanguage{russian}{арбитри}&\underline{\foreignlanguage{russian}{арбитри}}&\foreignlanguage{russian}{арбитрите}&definite form missing from targets\\
&revolt& \foreignlanguage{russian}{бунт}&\underline{\foreignlanguage{russian}{бунт}}&\foreignlanguage{russian}{бунта}&definite form missing from targets\\
& & \foreignlanguage{russian}{въстание} \\
&remembered& \foreignlanguage{russian}{запомнен}&\underline{\foreignlanguage{russian}{запомнен}}&\foreignlanguage{russian}{запомнена}&feminine form missing from targets\\
&hide& \foreignlanguage{russian}{скриване}&\foreignlanguage{russian}{скриване}&\foreignlanguage{russian}{скриват}&\textit{hide} as a verb vs. \textit{hide} as a noun\\
&bench& \foreignlanguage{russian}{пейката}&\foreignlanguage{russian}{пейка}&\foreignlanguage{russian}{скамейка}&synonym missing from targets\\
& & \foreignlanguage{russian}{пейка} \\
&depot& \foreignlanguage{russian}{депо}&\foreignlanguage{russian}{депо}&\textcolor{gray}{\foreignlanguage{russian}{гара}}&VM-S predicted `station'\\
&gaelic& \foreignlanguage{russian}{келтски}&\foreignlanguage{russian}{келтски}&\textcolor{gray}{\foreignlanguage{russian}{ирландският}}&VM-S predicted `the irish'\\
&footage& \foreignlanguage{russian}{кадри}&\foreignlanguage{russian}{кадри}&\textcolor{gray}{\foreignlanguage{russian}{заснети}}&VM-S predicted `shot'\\
\midrule
  \parbox[t]{2mm}{\multirow{14}{*}{\rotatebox[origin=c]{90}{VM-S \cmark, RCSLS \xmark }}}&egg& \foreignlanguage{russian}{яйцето}&\foreignlanguage{russian}{яйчен}&\foreignlanguage{russian}{яйце}&translation for attributive use of noun \\
& & \foreignlanguage{russian}{яйца}&&&missing from targets \\
& & \foreignlanguage{russian}{яйце} \\
&crowned& \foreignlanguage{russian}{коронован}&\foreignlanguage{russian}{коронована}&\underline{\foreignlanguage{russian}{коронован}}&feminine form missing from targets\\
&volcanic& \foreignlanguage{russian}{вулканична}&\underline{\foreignlanguage{russian}{вулканичен}}&\foreignlanguage{russian}{вулканична}&masculine form missing from targets\\
&penny& \foreignlanguage{russian}{пени}&\foreignlanguage{russian}{паричка}&\foreignlanguage{russian}{пени}&synonym missing from targets\\
&pound& \foreignlanguage{russian}{паунд}&\textcolor{gray}{\foreignlanguage{russian}{кило}}&\foreignlanguage{russian}{паунд}&RCSLS predicted a non-word\\
& & \foreignlanguage{russian}{кг} \\
&thursday& \foreignlanguage{russian}{четвъртък}&\textcolor{gray}{\foreignlanguage{russian}{петък}}&\foreignlanguage{russian}{четвъртък}&RCSLS predicted `friday'\\
&striker& \foreignlanguage{russian}{нападател}&\textcolor{gray}{\foreignlanguage{russian}{защитник}}&\foreignlanguage{russian}{нападател}&RCSLS predicted `defender'\\
& & \foreignlanguage{russian}{страйкър} \\
&pond& \foreignlanguage{russian}{езерце}&\textcolor{gray}{\foreignlanguage{russian}{къщичка}}&\foreignlanguage{russian}{езерце}&RCSLS predicted `cottage'\\
&flute& \foreignlanguage{russian}{флейтата}&\textcolor{gray}{\foreignlanguage{russian}{тромпет}}&\foreignlanguage{russian}{флейта}&RCSLS predicted `trumpet'\\
& & \foreignlanguage{russian}{флейта} \\
\midrule
  \parbox[t]{2mm}{\multirow{15}{*}{\rotatebox[origin=c]{90}{VM-S \xmark,   RCSLS \xmark }}}&circular& \foreignlanguage{russian}{кръгло}&\foreignlanguage{russian}{кръгла}&\foreignlanguage{russian}{кръгла}&feminine form missing from targets\\
&sailed& \foreignlanguage{russian}{отплава}&\foreignlanguage{russian}{отплавал}&\foreignlanguage{russian}{отплавал}&participle form missing from targets\\
&grants& \foreignlanguage{russian}{субсидии}&\foreignlanguage{russian}{стипендии}&\foreignlanguage{russian}{стипендии}&synonym missing from targets\\
&spots& \foreignlanguage{russian}{петна}&\foreignlanguage{russian}{петната}&\foreignlanguage{russian}{петната}&definite form missing from targets\\
&armies& \foreignlanguage{russian}{армии}&\foreignlanguage{russian}{армиите}&\foreignlanguage{russian}{армиите}&definite form missing from targets\\
&nose& \foreignlanguage{russian}{нос}&\textcolor{gray}{\foreignlanguage{russian}{врат}}&\textcolor{gray}{\foreignlanguage{russian}{задницата}}& RCSLS predicted `neck',\\
& & \foreignlanguage{russian}{носа} &&&VM-S predicted `bottom' \\
& & \foreignlanguage{russian}{носът} \\
&foods& \foreignlanguage{russian}{храни}&\textcolor{gray}{\foreignlanguage{russian}{сладкиши}}&\textcolor{gray}{\foreignlanguage{russian}{напитки}}&RCSLS predicted `sweets',\\
&&&&&VM-S predicted 'drinks'\\
&cliff& \foreignlanguage{russian}{скала}&\textcolor{gray}{\foreignlanguage{russian}{терас}}&\foreignlanguage{russian}{скалата}& RCSLS predicted non-word,\\
& & \foreignlanguage{russian}{клиф} &&&definite form missing from targets\\
&elevated& \foreignlanguage{russian}{повишени}&\textcolor{gray}{\foreignlanguage{russian}{понижен}}&\textcolor{gray}{\foreignlanguage{russian}{понижен}}&models predicted `reduced'\\
& & \foreignlanguage{russian}{повишена} \\
& & \foreignlanguage{russian}{повишен} \\
\end{tabular}
\end{adjustbox}
\caption{Example translations from {\sc en} to {\sc bg}. 
In cases where both models predicted forms of the same word, one being more canonical than the other, we underline the canonical form. Truly incorrect translations are marked in grey. Notice the high number of correct translations that are not listed as gold-standard targets.}
\label{tab:examples}
\end{table*}

\clearpage
\section*{D \space \space Appendix }
\setcounter{table}{5}

Table~6 shows an example of an inflectional correspondence map. It signifies that whenever an English word is encountered which is a verb in the infinitive, seven Bulgarian forms would be added to the list of targets, if not in it already. Addition of targets is also conditioned on their presence in the pretrained embeddings vocabulary.  

\begin{table}[b]
    \centering
    \resizebox{0.6\linewidth}{!}{
    \begin{tabular}{llccc}
         {\sc src} & {\sc tgt}\\
         \midrule
         V;NINF & V;IMP;2;SG\\
         & V;IMP;2;PL\\
         & V;IND;PRS;1;SG\\
         & V;IND;PRS;1;PL\\
         & V;IND;PRS;2;SG\\
         & V;IND;PRS;2;PL\\
         & V;IND;PRS;3;PL
         
    \end{tabular}
    }
    \caption{Example of an inflectional correspondence map.}
    \label{tab:corresp}
\end{table}

The modifications performed in this manner narrowed the gap in performance between RCSLS and VM-S by only 0.1 percentage points for {\sc en--bg} (from 6.7\% to 6.4\%) and by 1.6 percentage points for  {\sc en--de} (from 6.5\% to 4.9\%). Detailed results can be found in Table~7. Recall that for Bulgarian, we estimated 54\% of the gap in performance to stem from false False Positives. If the enrichment procedure was perfect, it should have reduced the gap from 6.6\% to less than 3.3\%. Unfortunately, due to limited coverage of the inflectional tables and of the pretrained embeddings, only 240 additional word forms were added to the {\sc en--bg} dictionary, making for a an almost negligible effect on precision. 

\definecolor{LightCyan}{rgb}{0.5,1,1}
\begin{table}
    \centering
    \begin{tabular}{ccc}
                            &                    {\sc de} & {\sc bg} \\
                        \rowcolor{LightCyan}
                        \multirow{2}{*}{VM-S}  & 65.5 & 49.8 \\
                                               & 67.6 & 50.3 \\
                        \rowcolor{LightCyan}
                        \multirow{2}{*}{RCSLS} & 72.0 & 56.4\\
                                               & 72.5  & 56.8\\
                        \rowcolor{LightCyan}
                        \multirow{2}{*}{$\Delta$} & 6.5 & 6.7\\
                                                  & 4.9  & 6.5
    \end{tabular}
    \caption{Results before (cyan rows) and after (white rows) coverage enrichment for {\sc de} and {\sc bg}}
    \label{tab:enrich}
\end{table}

\end{document}